\documentclass[a4paper,conference]{IEEEtran}

\usepackage{cite}
\usepackage{graphicx,color}
\usepackage{amsmath}

\usepackage[linesnumbered,ruled]{algorithm2e}
\usepackage{mathtools}
\usepackage{multirow}
\DeclareMathAlphabet      {\mathbfit}{OML}{cmm}{b}{it}

\begin{document}
%
\title{Domain Generalized Person Re-Identification\\
via Cross-Domain Episodic Learning}

\author{\IEEEauthorblockN{Ci-Siang Lin}
\IEEEauthorblockA{National Taiwan University\\
Taipei, Taiwan\\
ASUS Intelligent Cloud Services\\
Taipei, Taiwan\\
Email: d08942011@ntu.edu.tw}
\and
\IEEEauthorblockN{Yuan-Chia Cheng}
\IEEEauthorblockA{National Taiwan University\\
Taipei, Taiwan\\
Email: r08942154@ntu.edu.tw}
\and
\IEEEauthorblockN{Yu-Chiang Frank Wang}
\IEEEauthorblockA{National Taiwan University\\
Taipei, Taiwan\\
ASUS Intelligent Cloud Services\\
Taipei, Taiwan\\
Email: ycwang@ntu.edu.tw}}

\maketitle

\begin{abstract}
Aiming at recognizing images of the same person across distinct camera views, person re-identification (re-ID) has been among active research topics in computer vision. Most existing re-ID works require collection of a large amount of labeled image data from the scenes of interest. When the data to be recognized are different from the source-domain training ones, a number of domain adaptation approaches have been proposed. Nevertheless, one still needs to collect labeled or unlabelled target-domain data during training. In this paper, we tackle an even more challenging and practical setting, \textit{domain generalized (DG) person re-ID}. That is, while a number of labeled source-domain datasets are available, we do \textit{not} have access to any target-domain training data. In order to learn domain-invariant features without knowing the target domain of interest, we present an episodic learning scheme which advances meta learning strategies to exploit the observed source-domain labeled data. The learned features would exhibit sufficient domain-invariant properties while not overfitting the source-domain data or ID labels. Our experiments on four benchmark datasets confirm the superiority of our method over the state-of-the-arts.
\end{abstract}


%
\IEEEpeerreviewmaketitle

\section{Introduction}
\label{sec:intro}

Person re-identification (re-ID) \cite{zheng2016person} has been among active research topics in computer vision
due to its wide applications to person tracking \cite{andriluka2008people}, video surveillance systems \cite{khan2016person} and smart cities. Given a query image containing a person of interest, re-ID aims at matching gallery images with the same identity across different camera views. 
A number of works \cite{hermans2017defense, sun2019dissecting, wang2018learning, ge2018fd} have been proposed to recognize the identical identity suffering from the variation of viewpoints, postures, occlusions or background clutters. However, most of these approaches require collection of a large amount of labeled image data from the scenes of interest, which is not practical for real-world applications due to limited resources and privacy issues.  

Alternatively, one can utilize labeled data from one or multiple source domains, and jointly observe labeled or unlabeled target-domain data to train the re-ID models. Such domain adaptation (DA) approaches \cite{fu2019self} have been proposed for cross-dataset re-ID. If not observing label information from target-domain data during training, the resulting unsupervised domain adaptation (UDA) setting \cite{zhong2018generalizing, Liu_2019_CVPR, chen2019instance} would be a more difficult task to handle. Nevertheless, for cross-dataset re-ID, one still needs to collect target-domain data for training purposes.
\begin{figure}[t]
    \includegraphics[width=1.0\linewidth]{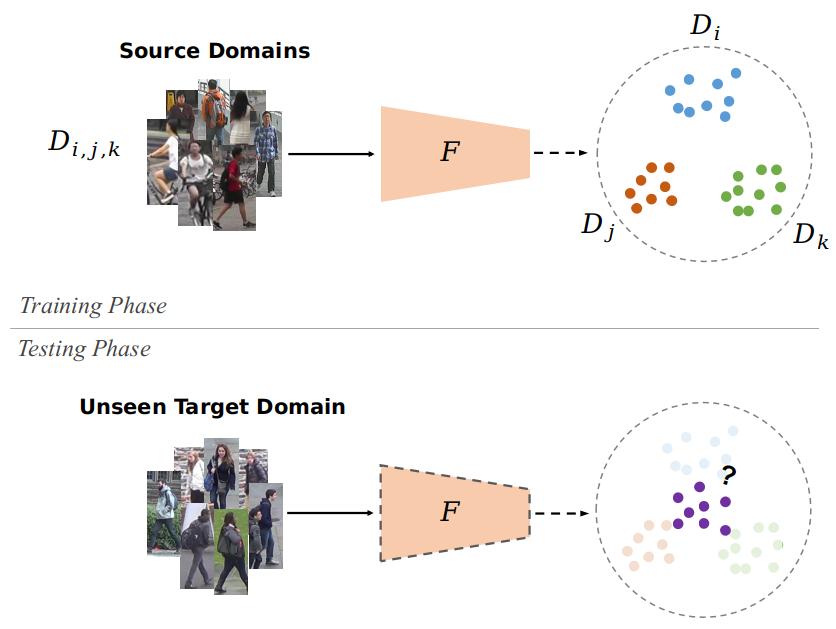}
    \caption{Illustration of domain generalized person re-identification. During training, multiple source-domain data are observed to learn domain-invariant representations, with the goal of tackling re-ID tasks in unseen target domains.}
    \label{fig:teasor}
\end{figure}
In this paper, we tackle an even more challenging and practical setting, \textit{domain generalized (DG) person re-ID}. As illustrated in Fig. \ref{fig:teasor}, while a number of labeled source-domain datasets are available, we do \textit{not} have access to the target-domain data even during training. Different from cross-dataset re-ID, the goal of domain generalized person re-ID is to improve the generalization and robustness of the learned model, which is learned from multiple source domains only. While a number of works \cite{ghifary2015domain, li2018learning, balaji2018metareg, Li_2019_ICCV} have been proposed for solving DG classification tasks, the label sets across domains (including the unseen target domain) remain the same. This is very different from the setting for re-ID, in which we do \textit{not} assume that persons of interest across domains remain the same. As a result, existing DG classification methods cannot easily tackle the re-ID tasks. For DG person re-ID, recent works like DualNorm \cite{jia2019frustratingly} takes advantage of Batch Normalization (BN) \cite{ioffe2015batch} and Instance Normalization (IN) \cite{ulyanov2016instance} to alleviate the domain difference, while DIMN \cite{Song_2019_CVPR} learns the mapping between person images and the ID classifiers but does not extended to unseen target-domain data well.

To address the challenging DG re-ID tasks without observing target-domain training data, we present an episodic learning scheme which advances meta learning strategies to exploit the observed source-domain labeled data. The learned features would exhibit sufficient domain-invariant properties while not overfitting the source-domain data or ID labels. This arms us to apply the learned model to any target domains of interest in no need of extra data collection and model updating. Compared to prior works, our experiments confirm that our proposed framework indeed improve the performance under the practically favorable setting.

We now highlight the contributions of our work below:\\

\begin{itemize}

\item We are among the first to derive domain invariant yet identity-discriminative features for re-ID without observing target-domain data during training.\\

\item We advance meta learning strategies, allowing derivation of domain-invariant latent representation with re-ID  guarantees.\\

\item Experimental results on four benchmark datasets quantitatively verify that our approach performs preferably against state-of-the-art (cross-dataset) re-ID methods.\\

\end{itemize}

\section{Related Works}
\label{sec:related}

\subsection{Person Re-Identification}

Person re-identification (re-ID)~\cite{zheng2016person} has been widely studied in the literature. With the advancement of deep learning, supervised person re-ID~\cite{hermans2017defense,sun2019dissecting,ge2018fd,tian2018eliminating,he2019foreground} has achieved a great progress in the last decade. Existing methods typically focus on tackling the challenges of matching images with viewpoint~\cite{sun2019dissecting} and pose variations~\cite{ge2018fd}, or those with background clutter~\cite{tian2018eliminating} or occlusion presented~\cite{he2019foreground}. For example, Sun et al.~\cite{sun2019dissecting} comprehensively analyze the influence of viewpoint on re-ID by varying the rotation angle of the pedestrian relative to the camera. With the guidance of human pose maps, Ge et al.~\cite{ge2018fd} develop a pose transferable GAN and derive pose invariant representations to handle pose variants. To mitigate the influence of background clutter, Tian et al.~\cite{tian2018eliminating} apply human parsing models to extract informative features of foreground regions. With no need for part-level alignment, He et al.~\cite{he2019foreground} construct spatial image pyramids that can re-identify persons accurately in the presence of heavy occlusion. While promising results have been observed, the above approaches typically requires a large amount of labeled data, thus limit themselves by scalability and practicality.

\subsection{Cross-Dataset Person Re-Identification}

Alternatively, one can utilize labeled data from one or multiple source domains, and jointly observed unlabeled target-domain data to train the re-ID models. Such approaches \cite{bak2018domain,deng2018image,chen2019instance,lin2018multi,wang2018transferable,zhong2018generalizing} aims to transfer and adapt identity-discriminative knowledge from labeled source domain to unlabeled targert domain data. Most of the works fall into three categories: (1) image-level style transfer (2) feature-level distribution alignment (3) joint image-level and feature-level alignment. The first category~\cite{bak2018domain,deng2018image,chen2019instance} typically performs image-image translation while preserving identity information based on CycleGAN~\cite{zhu2017unpaired}. The second one~\cite{lin2018multi,wang2018transferable} employs constraints like Maximum Mean Discrepancy (MMD)~\cite{lin2018multi} to derive a domain-invariant latent space. The last~\cite{zhong2018generalizing} achieves latent space and pixel space alignment jointly by combining techniques from both categories. While the aforementioned works are effective for cross-dataset person re-identification, one still needs to collect target-domain data for training purpose. 

\subsection{Person Re-Identification in Unseen Domains}

Learning from data across multiple source domains for handling the associated task in unseen domains is referred to as domain generalization. To tackle domain generalization, a plethora of methods \cite{blanchard2011generalizing, muandet2013domain, li2017deeper, li2018domain} have been proposed. Xu et al.~\cite{xu2014exploiting} introduces a low-rank structure to derive the corresponding feature representation by exemplar-SVMs. \cite{muandet2013domain, li2018domain} intends to learn a domain-invariant feature space by exploiting multiple source domains. Deemed as a mix-up of above, Li et al.~\cite{li2017deeper} decomposes the model into domain-specific and domain-invariant components and utilizes both to make predictions. Moreover, applying the meta-learning strategy has become a prevalent branch to address domain generalization. MLDG \cite{li2018learning} carries out modification on MAML~\cite{finn2017model} to adapt the learning scheme to domain generalization settings. Li et al.~\cite{li2019episodic} introduces an episodic training framework in which domain-specific feature extractors and classifiers are crossly trained in order to simulate interacting with a domain-specific tuned partner so that the feature extractors are able to learn robust feature representation. 
Despite the effectiveness of the above works, they cannot be easily extended to tackle re-id tasks since the setting of domain generalization typically assumes the label spaces are shared across domains (including the target one).

\begin{figure*}[t]
    \centering
    \includegraphics[width=1.0\linewidth]{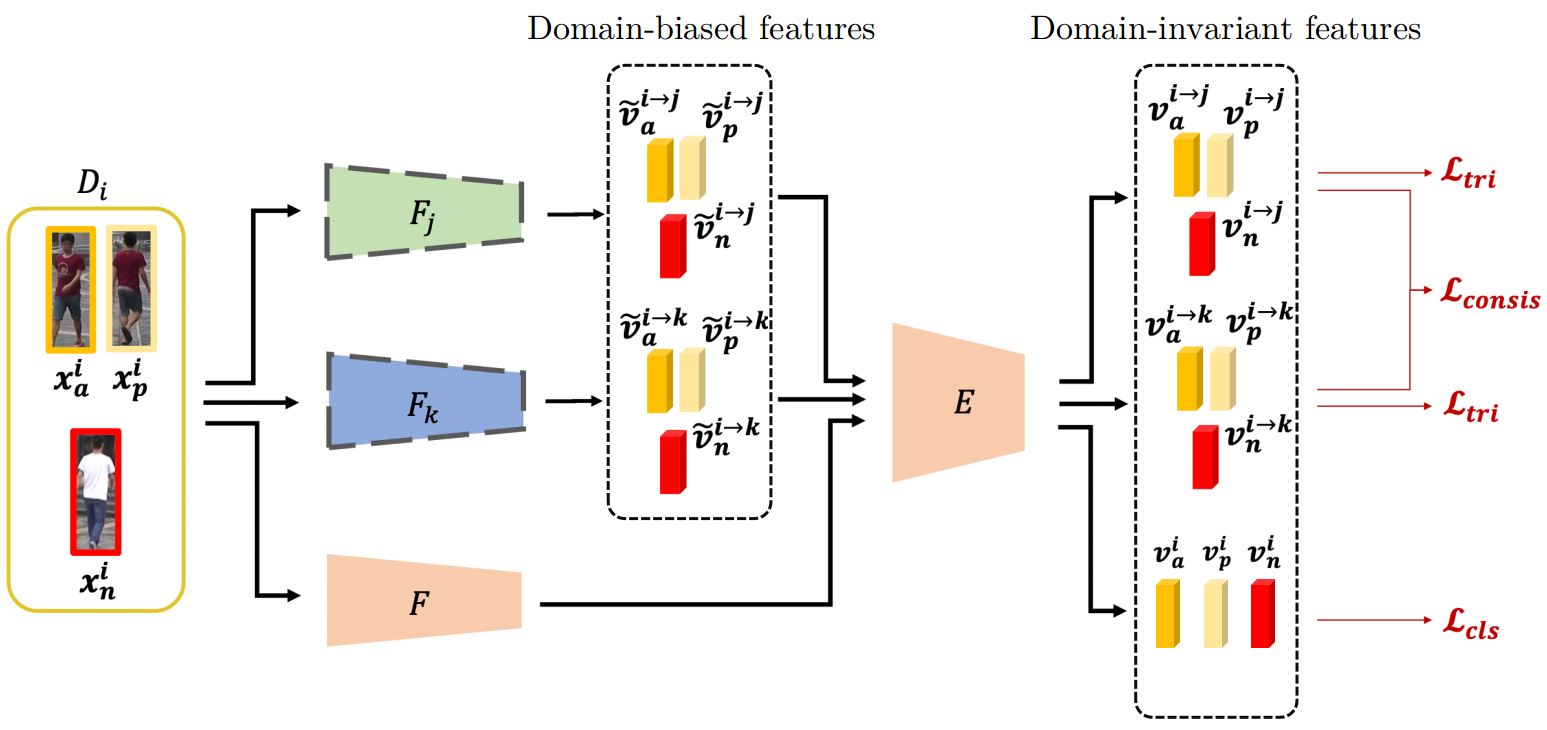}
    \caption{Overview of our proposed framework. During training, triplet images (i.e., anchor $x_a^i$, and its positive $x_p^i$ and negative neighbor $x_n^i$) from source domain $D_i$ are forwarded through feature extractors $F_{j}$, $F_{k}$ pretrained on domains $D_j$ and $D_k$, also through a global feature extractor $F$. The features derived by pretrained extractors are viewed as domain-biased features $\tilde{v}$, and the global encoder $E$ is expected to observe the resulting triplet loss $\mathcal{L}_{tri}$ and consistency loss $\mathcal{L}_{consis}$. The features derived by $F$ are forwarded through $E$ to guide classification losses $\mathcal{L}_{cls}$. During testing, only $F$ and $E$ are needed to extract as domain-invariant features $v$ for performing re-ID.}
    \label{fig:model}
\end{figure*}

Recent works have emerged to deal with re-ID under domain generalization settings. That is, one trains the model with no access to the data from target domain and are required to recognize the unseen person of interest. DualNorm leverages both Batch Normalization (BN)~\cite{ioffe2015batch} and Instance Normalization (IN)~\cite{ulyanov2016instance} to reduce the domain shift. DIMN~\cite{Song_2019_CVPR} proposes to map gallery person images into classifier weights with the help of memory bank. Both of them do not simulate the unseen domain scenario with meta-learning. In this work, we also consider domain generalization for re-ID and propose an episodic meta learning strategy to derive domain-invariant features.

\section{Proposed Method}
\label{sec:method}

\subsection{Notations and Problem Formulation}
\label{ssec:notations}

We first define the notations to be used in this paper. For domain generalized person re-ID, assume that we have the access to data from $N_d$ labeled source domains (datasets), i.e., source-domain data $D_S = \{D_i\}_{i=1}^{N_d}$, and the $ith$ source domain $D_i$ contains a set of $N_i$ images $X_i = \{x_u^i\}_{u=1}^{N_i}$ with the associate label set $Y_i = \{y_u^i\}_{u=1}^{N_i}$, where $x_u^i \in R^{H \times W \times 3}$ and $y_u^i \in R $ denote the $uth$ image and its corresponding identity label from the $ith$ source domain $D_i$, respectively. Note that for any pair of source domains $D_i$ and $D_j$, we consider their ID labels are \textit{disjoint}. This unique property makes our domain generalized person re-ID challenging yet practical than prior domain adaptation or domain generalization ones (recall that prior DA or DG approaches consider a joint label space across domains). 

To achieve DG re-ID, we present an end-to-end meta learning framework, as illustrated in Fig.~\ref{fig:model}. Our framework aims at learning domain-invariant features $\mathbfit{v}$ by learning feature and re-ID encoders $F$ and $E$, based on feature extractors pretrained on each source domain. During each episode, three domains $D_i$, $D_j$ and $D_k$ are randomly sampled from the source domain datasets. Feature extractors $F_j$ and $F_k$ pretrained on $D_j$ and $D_k$ are used to extract the domain-biased features $\mathbfit{\tilde{v}}^{i \rightarrow j}$ and $\mathbfit{\tilde{v}}^{i \rightarrow k}$ for images ${x}^{i}$ from domain $D_i$, respectively. With the proposed meta learning scheme, our encoder $E$ would derive domain-invariant yet identity-discriminative features for re-ID purposes. 
The details of our proposed learning framework will be discussed in the following sub-sections.

\subsection{Meta Learning for Domain-Invariant Representation}
\label{ssec:invariant}

We now detail how we advance meta learning strategies for deriving domain-invariant features, without observing the target-domain data during training. Our proposed framework starts with pretrained domain-specific feature extractor $F_i$ using labeled data in each source domain $D_i$ (e.g., using TripletLoss \cite{hermans2017defense} as most re-ID works do). Thus, a total of $N_d$ pretrained domain-specific feature extractors are obtained and will later be utilized in our meta learning framework.

As shown in Fig~\ref{fig:model}, our goal is to learn a domain-invariant feature extractor $F$, followed by a domain generalized encoder $E$, for deriving DG re-ID. To accomplish this, we present an \textit{episodic learning} scheme that utilizes the pretrained $F_i$ for learning the above network modules. To be more precise, in each episode during training, we randomly select data from three source domains $D_i$, $D_j$ and $D_k$. For an input anchor image $x_a^i$ from domain $D_i$, we particularly apply the pretrained domain-specific feature extractors $F_j$ and $F_k$ to output the associated \textit{domain-biased} features $\mathbfit{\tilde{v}}_a^{i \rightarrow j}$ and $\mathbfit{\tilde{v}}_a^{i \rightarrow k}$. In other words, we have $\mathbfit{\tilde{v}}_a^{i \rightarrow j}=F_j(x^i_a)$ and $\mathbfit{\tilde{v}}_a^{i \rightarrow k}=F_k(x^i_a)$. Since $\mathbfit{\tilde{v}}_a^{i \rightarrow j}$ and $\mathbfit{\tilde{v}}_a^{i \rightarrow k}$ are both derived from the same image $x_a^i$, we require the domain generalized encoder $E$ to output the final \textit{domain-invariant} features $\mathbfit{v}_a^{i \rightarrow j}$ and $\mathbfit{v}_a^{i \rightarrow k}$. To enforce $E$ to preserve the domain-invariant yet re-ID preserved information from $\mathbfit{v}_a^{i \rightarrow j}$ and $\mathbfit{v}_a^{i \rightarrow k}$, we propose to calculate the \emph{cross-domain consistency loss} $\mathcal{L}_{consis}$ on the above feature pair:
\begin{equation}
  \label{eq:consis}
  \begin{aligned}
  \mathcal{L}_\mathrm{consis} = &~ {E}_{x_a^i \sim X_i}\|\mathbfit{v}_a^{i \rightarrow j} - \mathbfit{v}_a^{i \rightarrow k}\|_2.
  \end{aligned}
\end{equation}
Note that the pretrained feature extractors $F_j$ and $F_k$ are fixed, which would \emph{not} be updated by back-propagated gradients calculated from $\mathcal{L}_{consis}$. This is the \textit{key} technique for each domain-specific feature extractors to preserve its own domain-specific properties, while allowing the domain generalized encoder $E$ to extract domain-invariant features during the episodic learning process. It is worth repeating that, during learning of domain-invariant features, no target-domain data are observed.

We also note that, we choose \emph{not} to apply adversarial learning techniques  (e.g., DANN \cite{ganin2016domain}) for deriving domain-invariant features. This is because that, the person identities are \textit{disjoint} across source domains. If one applies adversarial learning or similar domain adaptation techniques for eliminating the domain differences, it is likely that the person ID or pose information is also confused by such learning strategy, which is not desirable for re-ID tasks.

\begin{algorithm}[t]

\textbf{Input}: $N_d$ source domains $D_S = \{D_i\}_{i=1}^{N_d}$

\BlankLine
\textbf{Stage 1: Pretraining}

\For{each domain $D_i$ in $D_S$}{
Pretrain a domain-specific feature extractor $F_i$ using data ($X_i$, $Y_i$) and TripletLoss
}

\BlankLine
\textbf{Stage 2: Episodic Training}

$\theta_{F}$, $\theta_{E}$ $\leftarrow$ initialize

\For{num. of iterations}{
$D_i$, $D_j$, $D_k$ $\leftarrow$ randomly sampled from $D_S$\\
$x_a^i$, $x_p^i$, $x_n^i$ $\leftarrow$ randomly sampled from $D_i$\\
$F_j$, $F_k$ $\leftarrow$ pretrained in domain $D_j$ and $D_k$\\
\BlankLine
$\mathbfit{\tilde{v}}_a^{i \rightarrow j}$, $\mathbfit{\tilde{v}}_a^{i \rightarrow k}$ $\leftarrow$ obtained from $F_j(x_a^i)$, $F_k(x_a^i)$

$\mathbfit{v}_a^{i \rightarrow j}$, $\mathbfit{v}_a^{i \rightarrow k}$ $\leftarrow$ obtained from $E(\mathbfit{\tilde{v}}_a^{i \rightarrow j})$, $E(\mathbfit{\tilde{v}}_a^{i \rightarrow k})$

$\mathcal{L}_{consis}$ $\leftarrow$ calculated by (\ref{eq:consis})
\BlankLine
$\mathbfit{\tilde{v}}_p^{i \rightarrow j}$, $\mathbfit{\tilde{v}}_n^{i \rightarrow j}$ $\leftarrow$ obtained from $F_j(x_p^i)$, $F_j(x_n^i)$

$\mathbfit{v}_p^{i \rightarrow j}$, $\mathbfit{v}_n^{i \rightarrow j}$ $\leftarrow$ obtained from $E(\mathbfit{\tilde{v}}_p^{i \rightarrow j})$, $E(\mathbfit{\tilde{v}}_n^{i \rightarrow j})$

$\mathcal{L}_{tri}$ $\leftarrow$ calculated by (\ref{eq:tri})

$\mathcal{L}_{cls}$ $\leftarrow$ calculated by (\ref{eq:cls})

\BlankLine
$\mathcal{L}_{total} = {L_{cls}} + \lambda_{tri}  \cdot {L_{tri}} + \lambda_{consis} \cdot {L_{consis}}$

$\theta_{E, F}	\leftarrow \theta_{E, F} - \eta \cdot \bigtriangledown L_{total}$

}

\textbf{Output}: Global feature extractor $F$ and domain generalized encoder $E$
\caption{Training of the Proposed Episodic Learning Scheme}
\label{alg:pdanet}
\normalsize
\end{algorithm}

\subsection{Domain Generalized Person Re-ID}
\label{ssec:dgreid}

To utilize label information observed from source-domain data for learning domain-invariant features for re-ID, we adopt the triplet loss on the derived domain invariant space (i.e., feature space output by domain generalized encoder $E$). That is, for each domain invariant feature $\mathbfit{v}_a^{i \rightarrow j}$ derived from the input anchor image $x_a^i$, a triplet tuple is composed of $\mathbfit{v}_p^{i \rightarrow j}$ with the same identity label as $x_a^i$ and $\mathbfit{v}_n^{i \rightarrow k}$ with different identity label as $x_a^i$. Then, the distances $d_\mathrm{p}$ and $d_\mathrm{n}$ for such positive and negative pairs are defined as:
\begin{equation}
  \begin{aligned}
  d_\mathrm{p} = \|\mathbfit{v}_a^{i \rightarrow j} - \mathbfit{v}_p^{i \rightarrow j}\|_2,
  \end{aligned}
  \label{eq:dp}
\end{equation}
\begin{equation}
  \begin{aligned}
  d_\mathrm{n} = \|\mathbfit{v}_a^{i \rightarrow j} - \mathbfit{v}_n^{i \rightarrow j}\|_2,
  \end{aligned}
  \label{eq:dn}
\end{equation}
where $\mathbfit{v}_a^{i \rightarrow j}$, $\mathbfit{v}_p^{i \rightarrow j}$ and $\mathbfit{v}_n^{i \rightarrow j}$ denote the domain-invariant latent features of $x_a^i$, $x_p^i$ and $x_n^i$ derived from the encoder $E$. With the above definitions, we have the \emph{domain generalized triplet loss}, $\mathcal{L}_{tri}$, which is calculated as:
\begin{equation}
  \begin{aligned}
  \mathcal{L}_{tri}
  = &~ {E}_{(x_a^i,y_a^i) \sim (X_i,Y_i)}\max(0, m + d_\mathrm{p} - d_\mathrm{n}),
  \end{aligned}
  \label{eq:tri}
\end{equation}
where $m > 0$ is the margin enforcing the separation between positive and negative image pairs. 

Take computational feasibility into consideration, we  follow \cite{hermans2017defense} and select only the hardest positive and negative samples to calculate the triplet loss. By minimizing $\mathcal{L}_{tri}$, the encoder $E$ manages to capture the identity information from domain specific features $\mathbfit{\tilde{v}}_a^{i \rightarrow j}$ to extract domain invariant but identity-discriminative features $\mathbfit{v}_a^{i \rightarrow j}$. It is worth repeating that, the pretrained domain specific feature extractors $F_j$ and $F_k$ will \textit{not} be updated by this loss. 

\begin{table}[t]
    \caption{Data Statistics of Source-Domain Datasets}
    \begin{center}
    \resizebox{1.0\linewidth}{!}{
    \begin{tabular}{c|c|c|c}
    \hline
    Dataset                                 & Cameras  & IDs  & Images\\
    \hline
    Market1501~\cite{zheng2015scalable}     & 6        & 1,501 & 29,419\\
    DukeMTMC-reID~\cite{zheng2017unlabeled} & 8        & 1,812 & 36,411\\
    CUHK02~\cite{li2013locally}             & 5        & 1,816 & 7,264\\
    CUHK03~\cite{li2014deepreid}            & 6        & 1,467 & 14,097\\
    \hline
    Total                                   & 25       & 6,596 & 87,191\\
    \hline
    \end{tabular}}
    \end{center}
    \label{tab:source}
\end{table}
\begin{table}[t]
    \caption{Data Statistics of Target-Domain Datasets}
    \begin{center}
    \resizebox{1.0\linewidth}{!}{
    \begin{tabular}{c|c|c|c|c|c}
     \hline
     Dataset                             & Pr. IDs & Ga. IDs & Pr. Images & Ga. Images \\
     \hline
     GRID~\cite{loy2009multi}            & 125 & 900 & 125 & 1,025\\
     i-LIDS~\cite{Zheng2009AssociatingGO}& 60  & 60  & 60  & 60\\
     PRID~\cite{hirzer2011person}        & 100 & 649 & 100 & 649\\
     VIPeR~\cite{gray2008viewpoint}      & 316 & 316 & 316 & 316\\
     \hline
    \end{tabular}}
    \end{center}
    \label{tab:target}
\end{table}

To make the training process more stable and have the encoder $E$ describe global identity features, we additionally train a global (or domain-invariant) feature extractor $F$ using training data from \textit{all} source domains (datasets). To further enforce the re-ID capability, an additional classifier $C$ is integrated into the encoder $E$ for identity prediction, with the cross-entropy loss calculated as:
\begin{equation}
  \begin{aligned}
  \mathcal{L}_{cls} = &~ {E}_{(x, y) \sim D_S}-\log p(y|x),
  \end{aligned}
  \label{eq:cls}
\end{equation}
where $p(y|x)$ denotes the prediction probability indicating the image $x$ belongs to its corresponding identity label $y$.

With the above meta learning framework together with the introduced losses, the total loss $\mathcal{L}_{total}$ of our model is
\begin{equation}
  \begin{aligned}
  \mathcal{L}_{total} = &~ {L_{cls}}+ \lambda_{tri}  \cdot {L_{tri}} + \lambda_{consis}  \cdot {L_{consis}},
  \end{aligned}
  \label{eq:total}
\end{equation}
where $\lambda_{tri}$ and $\lambda_{consis}$ are the hyperparameters. To perform person re-ID on the unseen domain in the testing phase, the query and gallery images are forwarded through feature extractor $F$ and the encoder $E$ to derive domain-invariant re-ID features, which are applied for matching query/gallery images via nearest neighbor search in Euclidean distances.

\section{Experiments}
\label{sec:exp}

\subsection{Datasets and Experimental Settings}
\label{ssec:data}

To evaluate our proposed method, following \cite{Song_2019_CVPR} we use existing large-scale re-ID datasts as the source domains and test the performance on several target datasets which are not observed during training. To be specific, the source domains include Market1501 \cite{zheng2015scalable}, DukeMTMC-reID \cite{zheng2017unlabeled}, CUHK02 \cite{li2013locally} and CUHK03  \cite{li2014deepreid}, with a total of 6596 identities and 87191 images. The target datasets include GRID \cite{loy2009multi}, i-LIDS \cite{Zheng2009AssociatingGO}, PRID \cite{hirzer2011person} and VIPeR \cite{gray2008viewpoint}. We follow the single-shot setting with the number of probe/gallery images set as: GRID: 125/1025; i-LIDS: 60/60; PRID 100/649; VIPeR 316/316 respectively. Detailed data statistics are summarized in Table~\ref{tab:source} and~\ref{tab:target}. The average rank-1 accuracy over 10 random splits is reported based on the standard evaluation protocol and the cumulative matching curve (CMC). To be clear, we reproduce all methods and performances due to the lack of legal access to certain datasets used in original papers.

\subsection{Implementation Details}
\label{ssec:impl}
We implement our method using PyTorch. We use the backbone model proposed in \cite{jia2019frustratingly} as our global feature extractor $F$ and use ResNet-$50$ \cite{he2016deep} pretrained on ImageNet for domain specific modules $F_i$. The global encoder $E$ consists of two fully-connected (FC) layers with BatchNorm \cite{ioffe2015batch} layers while the global classifier $C$ is a single FC layer. We resize all the input images to $256 \times 128 \times 3$ (denoting height, width and channel, respectively). Random clipping and cropping are adapted for data augmentation. The margin $m$ is set as $0.3$ and we fix $\lambda_{tri}$ and $\lambda_{consis}$ as $0.2$ and $0.01$, respectively. We train our model for 150 epochs with the SGD optimizer. The initial learning rate is set as 0.01 and is decreased to 0.001 at 100 epochs. Label smoothing is used to prevent overfitting.

\subsection{Comparisons Against State-of-the-Art }
\label{ssec:sota}

\begin{table}[t]
    \caption{Performance comparisons of domain generalization based methods in terms of averaged Rank-1 accuracy. Note that target-domain data are only seen during testing.}

    \begin{center}
    \resizebox{1.0\linewidth}{!}{
    \begin{tabular}{c|c|c|c|c|c}
     \hline
     Target & GRID & i-LIDS & PRID & VIPeR & Avg. \\
     \hline
     DIMN~\cite{Song_2019_CVPR} & 23.4 & 44.8 & 13.1 & 29.9 & 27.8 \\
     DualNorm~\cite{jia2019frustratingly} & 29.2 & 58.3 & 54.3 & {\bf 38.6} & 45.1 \\
     \hline
     Ours~ & {\bf 33.0} & {\bf 62.3} & {\bf 57.6} & 38.5 & {\bf 47.8} \\
     \hline
    \end{tabular}
    }
    \end{center}
    
    \label{tab:sota}
\end{table}
\begin{table}[t]
    \caption{Performance comparisons of domain adaptation based methods in terms of averaged Rank-1 accuracy. Note that baseline and DANN observe only source-domain data during training without the access to any information from target domain.}

    \begin{center}
    \resizebox{1.0\linewidth}{!}{
    \begin{tabular}{c|c|c|c|c|c}
     \hline
     Target & GRID & i-LIDS & PRID & VIPeR & Avg. \\
     \hline
     Baseline~ & 18.8 & 52.5 & 14.8 & 32.0 & 29.5 \\
     DANN~\cite{ganin2016domain} & 29.0 & 57.2 & 56.8 & 37.8 & 45.2 \\
     \hline
     Ours & {\bf 33.0} & {\bf 62.3} & {\bf 57.6} & {\bf 38.5} & {\bf 47.8} \\
     \hline
    \end{tabular}
    }
    \end{center}

    \label{tab:dann}
\end{table}
\begin{table}[t]
    \caption{Ablation studies analyzing the importance of each introduced loss function.}

    \begin{center}
    \resizebox{1.0\linewidth}{!}{
    \begin{tabular}{c|c|c|c|c|c}
     \hline
     Target & GRID & i-LIDS & PRID & VIPeR & Avg. \\
     \hline
     Ours w/o $\mathcal{L}_{tri}$ & 31.3 & 59.0 & 55.8 & 37.3 & 45.8 \\
     Ours w/o $\mathcal{L}_{consis}$ & 30.6 & 60.3 & 55.7 & {\bf 40.1} & 46.7 \\
     \hline
     Ours~ & {\bf 33.0} & {\bf 62.3} & {\bf 57.6} & 38.5 & {\bf 47.8} \\
     \hline
    \end{tabular}
    }
    \end{center}

    \label{tab:abl}
\end{table}

In Table \ref{tab:sota}, we compare our proposed method with two state-of-the-arts \cite{jia2019frustratingly, Song_2019_CVPR} which attempted the DG setting for re-ID. From this table, we see that our method performed favorably well and observed performance margins over the state-of-the-art methods. We achieved the averaged~\textbf{Rank-1 accuracy} of~\textbf{47.8\%} on the four target datasets. Compared to DIMN \cite{Song_2019_CVPR}, our method learned domain invariant representations, while DMIN learned a mapping between a person image and its identity classifier weight without deriving a domain invariant latent space, and thus failed to generalize to target domain. Compared to DualNorm \cite{jia2019frustratingly} which simply adopted BN and IN layers to alleviate the domain shift, our method meta-learned to capture identity information under the proposed episodic scheme, and thus our averaged Rank-1 accuracy was higher by~\textbf{2.7\%}. From the experiment, the effectiveness of our model for domain generalized person re-ID was quantitatively verified.

As mentioned in Sec. \ref{ssec:invariant}, we did \emph{not} apply the adversarial training strategy to derive domain invariant representations since person identities are disjoint for any two different re-ID datasets, and thus the adversarial training strategy might produce pose-invariant or camera-invariant features instead of domain-invariant ones. To verify this, we compare our method to a simple baseline and DANN \cite{ganin2016domain}, which derived domain invariant features via adversarial loss and is a popular UDA method. For the baseline, we simply trained a single ResNet-$50$ to predict the person identities for all domains. Although DANN was originally proposed for UDA, we repurposed it for domain generalization by adding an auxiliary classifier to the baseline model with a gradient-reversal layer for domain confusion. As shown in Table \ref{tab:dann}, our averaged Rank-1 accuracy was higher than DANN by~\textbf{2.6\%}. This demonstrated our cross-domain consistency loss is practically more preferable than the adversarial loss.

\subsection{Ablation Studies}
\label{ssec:abl}

To further analyze the importance of each introduced loss function, we conduct ablation studies shown in Table \ref{tab:abl}. Without the domain generalized triplet loss $\mathcal{L}_{tri}$, our model would not properly capture the identity information for unseen domain, resulting in $2\%$ performance drop. Also, when $\mathcal{L}_{consis}$ is turned off, our model would fail to generalize to unseen domain since there is no explicit constraint for learning domain invariant representations. From the above experiment,  we confirmed that each introduced loss function is vital and beneficial to domain generalized person re-ID.
\section{Conclusions}
\label{sec:concl}

In this paper, we addressed the challenging domain generalized person re-ID problem, in which target-domain data is not available during training. With the proposed meta learning framework, we utilized episodic training strategy with pretrained domain specific feature extractors, and learn domain-invariant yet identity-discriminative features with re-ID performance guarantees. Our experiments on multiple benchmark datasets confirmed that our approach performed favorably against state-of-the-art domain adaptation and domain generalization methods on this challenge task.
\section*{Acknowledgment}
This work is supported in part by the Ministry of Science and Technology of Taiwan under grant MOST 109-2634-F-002-037.






%
\bibliographystyle{IEEEtran}
\bibliography{IEEEabrv, ref}

\end{document}